\documentclass[letterpaper, 10 pt, conference]{ieeeconf}  

\IEEEoverridecommandlockouts                              

\usepackage{graphicx}
\usepackage{graphics}
\usepackage{amsmath} 
\usepackage{amssymb}  
\usepackage{tabularx,longtable,multirow}
\usepackage{color}

\title{\LARGE \bf
Real-time 3D Shape Instantiation for Partially-deployed Stent Segment from a Single 2D Fluoroscopic Image in Robot-assisted Fenestrated Endovascular Aortic Repair
}

\author{Jian-Qing Zheng$^{*1}$, Xiao-Yun Zhou$^{*1}$ and Guang-Zhong Yang$^{1}$
\thanks{*Jian-Qing Zheng and Xiao-Yun Zhou contribute equally to this paper.}
\thanks{This work was supported by Engineering and Physical Sciences Research Council (EPSRC) project grant EP/L020688/1.}
\thanks{$^{1}$Jian-Qing Zheng, Xiao-Yun Zhou, and Guang-Zhong Yang are with the Hamlyn Centre for Robotic Surgery, Imperial College London, UK.
        {\tt\small j.zheng17@imperial.ac.uk}}%
}

\begin{document}

\maketitle
\thispagestyle{empty}
\pagestyle{empty}

\begin{abstract}
In robot-assisted Fenestrated Endovascular Aortic Repair (FEVAR), accurate alignment of stent graft fenestrations or scallops with aortic branches is essential for establishing complete blood flow perfusion. Current navigation is largely based on 2D fluoroscopic images, which lacks 3D anatomical information, thus causing longer operation time as well as high risks of radiation exposure. Previously, 3D shape instantiation frameworks for real-time 3D shape reconstruction of fully-deployed or fully-compressed stent graft from a single 2D fluoroscopic image have been proposed for 3D navigation in robot-assisted FEVAR. However, these methods could not instantiate partially-deployed stent segments, as the 3D marker references are unknown. In this paper, an adapted Graph Convolutional Network (GCN) is proposed to predict 3D marker references from 3D fully-deployed markers. As original GCN is for classification, in this paper, the coarsening layers are removed and the softmax function at the network end is replaced with linear mapping for the regression task. The derived 3D and the 2D marker references are used to instantiate partially-deployed stent segment shape with the existing 3D shape instantiation framework. Validations were performed on three commonly used stent grafts and five patient-specific 3D printed aortic aneurysm phantoms. Comparable performances with average mesh distance errors of 1$\sim$3mm and average angular errors around $7^\circ$ were achieved.

\end{abstract}

\section{Introduction}
\begin{figure}[th]
     \centering
      \framebox{\parbox{3.3in}{\includegraphics[width = 3.3in]{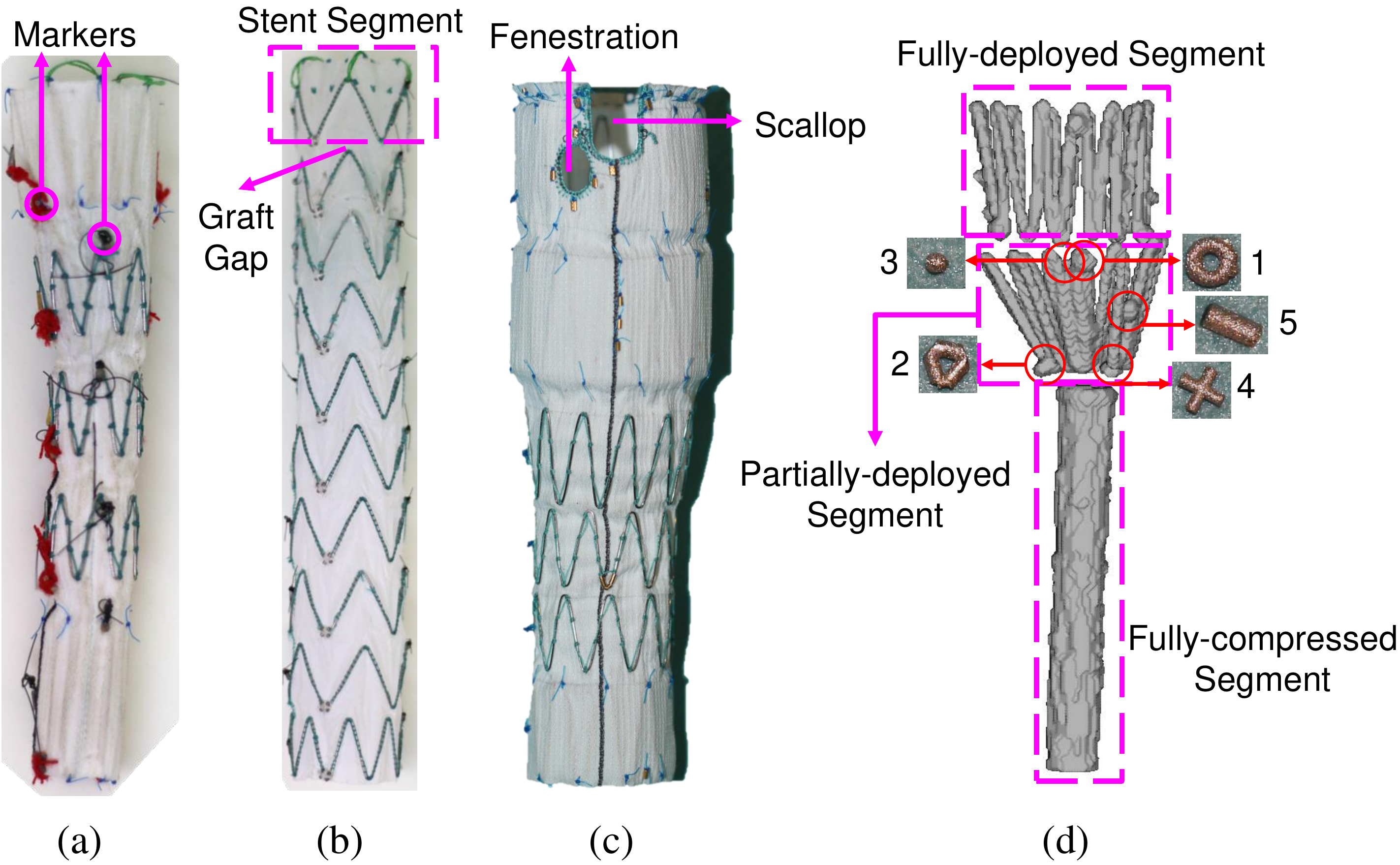}}}
      \caption{Illustration of iliac stent graft (a), thoracic stent graft (b), fenestrated stent graft (c), marker number and different stent segment status (d).}
      \label{fig:intro}
\end{figure}

Abdominal Aortic Aneurysm (AAA), an enlargement of the abdominal aorta with 50\% diameter over normal state, occurs increasingly often among old people \cite{sakalihasan2005abdominal}. The rupture of AAA brings in 85\%-90\% fatality rate \cite{kent2014abdominal}. Fenestrated Endovascular Aortic Repair (FEVAR) is a minimally invasive surgery for AAA, where a deployment catheter carrying a compressed stent graft is inserted via the femoral artery, advanced through the vasulature and deployed subsequently at the AAA position. Three typical stent grafts - iliac, fenestrated and thoracic stent graft are shown in Figure~\ref{fig:intro}(a), \ref{fig:intro}(b) and \ref{fig:intro}(c) respectively. In FEVAR, an accurate alignment of stent graft fenestrations or scallops (as shown in Figure~\ref{fig:intro}c) to aortic branches, i.e., renal arteries, is necessary for connecting branch stent grafts into aortic branches \cite{cross2012fenestrated}. Although several robot-assisted systems have been developed to facilitate the FEVAR procedure, i.e., the Magellan system (Hansen Medical, CA, USA), current navigation technique is still based on 2D fluoroscopic images which are insufficient for 3D-to-3D alignment. Either supplying 3D navigation for the AAA or fenestrated stent grafts would improve the navigation.

For 3D AAA navigation, a skeleton-based as-rigid-as-possible approach was proposed to adapt a 3D pre-operative AAA shape to intra-operative position of the deployment device from two fluoroscopic images for recovering the 3D AAA shape \cite{toth2015adaption}. A skeleton instantiation framework for AAA with a graph matching method and skeleton deformation was introduced to instantiate the 3D AAA skeleton from a single 2D fluoroscopic image \cite{zheng20183d}.

For offering 3D navigation for fenestrated stent grafts, many methods have been implemented. The 3D stent shape was recovered from a 2D X-ray image via registration and optimization in \cite{demirci20113d} but without estimation of the graft nor the angle or position of fenestrations or scallops. A 3D shape instantiation framework with stent graft modelling and Robust Perspective-n-Point (RPnP) method was proposed to instantiate the 3D shape of a fully-compressed stent graft \cite{zhoustent}. The work in \cite{zhoustent} was then used to recover the 3D shape of each stent segment (as shown in Figure~\ref{fig:intro}b), with customized markers, while Focal U-Net and graft gap (as shown in Figure~\ref{fig:intro}b) interpolation were proposed to semi-automatically segment customized markers and recover the whole 3D shape of fully-deployed stent grafts in \cite{zhou2018real_ral}. Equally-weighted Focal U-Net was also proposed for automatic marker segmentation in \cite{zhou2018towards_iros} to improve the automation of the 3D shape instantiation framework. However, the method by Zhou et al. could not instantiate the 3D shape of a partially-deployed stent segment, as the 3D marker references required by the RPnP method are unknown.

The method proposed in this paper aims to obtain the deformation pattern between partially-deployed and fully-deployed stent segment using deep learning based methods. General artificial neural networks can be applied to this task but with very large searching space of parameters. The relationship between each two markers is not uniform and the topological structure is non-Euclidean either. The classical convolutional kernel and thus the convolutional neural networks cannot be used for this problem. A novel convolution on an undirected simple graph called spectral graph convolution was described in \cite{shuman2012emerging}. A Graph Convolutional Network (GCN) with locally connected architecture was then proposed in \cite{bruna2013spectral} with $\mathcal{O}(n)$ parameter number for each layer based on the spectrum of graph Laplacian, which was validated on the MNIST dataset. Furthermore, a more efficient GCN with localized spectral convolution on a graph was proposed in \cite{defferrard2016convolutional}, reducing the parameter number to $\mathcal{O}(K)$ with improved performance on the MNIST dataset but computational complexity, where $K<n$ is the localized filter size. Another construction of GCN was also proposed in \cite{kipf2016semi} with first-order approximation of spectral graph convolutions for a large-scale architecture, but with less capacity for the same layer number compared to \cite{defferrard2016convolutional}.

\begin{figure}[th]
     \centering
      \framebox{\parbox{3.3in}{\includegraphics[width = 3.3in]{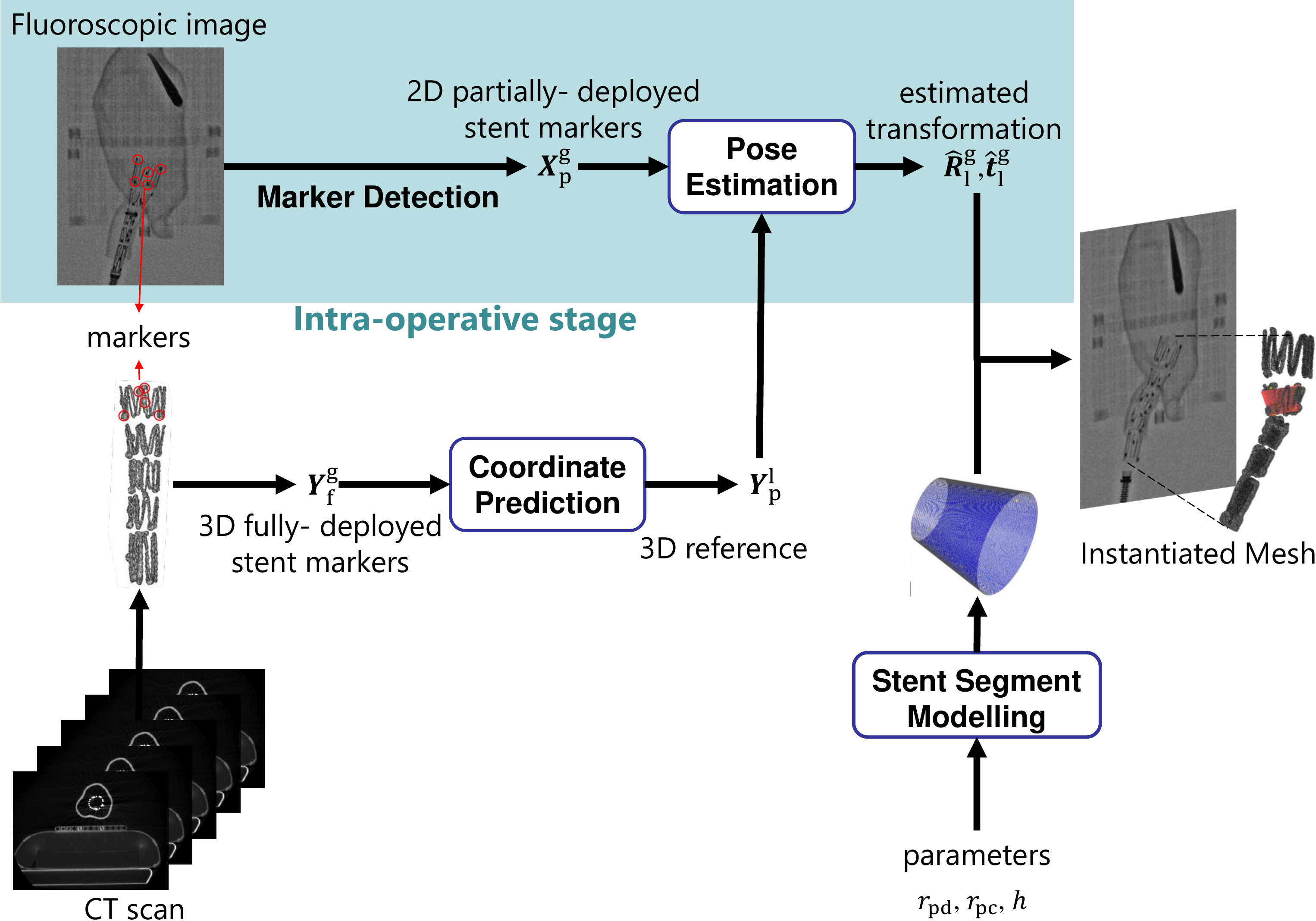}}}
      \caption{Pipeline for shape instantiation of partially-deployed stent segment from a single fluoroscopic image and the 3D CT scan of fully-deployed stent graft}
      \label{fig:pipeline}
\end{figure}

An Adapted GCN based on the architecture in \cite{defferrard2016convolutional} is proposed for predicting 3D marker references of partially-deployed stent segment from 3D fully-deployed markers, which bridges the gap of utilizing the RPnP method for 3D shape instantiation of partially-deployed stent segment. The coarsening layers are removed and the softmax function at the network end is replaced with a linear mapping. The derived 3D marker references are integrated into a previously deployed 3D shape instantiation framework \cite{zhou2018real_ral}, with the customized marker placement, stent segment modelling and the RPnP method, to achieve 3D shape instantiation for partially-deployed stent segment, The pipeline is shown in Figure~\ref{fig:pipeline}. Three stent grafts with total 26 different stent segments were used for the validation. Details regarding the methodology and experimental setup are in Section~\ref{sec:method}. Results with an average angular error about $7^\circ$ and an average mesh distance error around 2mm are stated in Section~\ref{sec:result}. Discussion and conclusion are introduced in Section~\ref{sec:discussion} and Section~\ref{sec:conclusion} respectively.

\section{Methodology}
\label{sec:method}
In this section, we introduce the proposed Adapted GCN for predicting 3D marker references, while briefly introducing the stent segment modelling and 3D shape instantiation to facilitate the understanding of the whole framework. Experimental setup is also demonstrated.

\subsection{Partially-deployed Stent Segment Modelling}
\label{sec:stent_model}
In practice, the parameters of stent segment, including the height and diameters at the fully-deployed and fully-compressed state, can be obtained via fenestrated stent graft and deployment catheter design. In this paper, as the stent grafts were experimented multiple times with compression and deployment, the practical parameters are different from the ideally designed ones and are measured manually.

In \cite{zhou2018real_ral}, a stent graft was modelled as a cylinder fitted by a series of concentric circles with a finite set of vertices $\mathcal{V}$ of coordinates $\textit{\textbf{V}}\in\mathbb{R}^{3\times(360h/0.1{\rm mm})}$. The coordinate of each circle vertex is defined as $(r {\rm cos}\theta~r {\rm sin}\theta~h)^\top$. In this paper, each partially-deployed stent segment is modelled as a cone with the diameters and the height of this segment.

Different from the fully-deployed stent segment in \cite{zhou2018real_ral}, the diameters of partially-deployed stent segments are not only decided by the designed deployed size but also the compression diameters $r_{\rm fc}\in\mathbb{R}_+$ and the gap width $w_{\rm g}\in\mathbb{R}_+$. In the experiments, one partially-deployed stent segment's diameter of its deployed side $r_{\rm pd}\in\mathbb{R}_+$ is set as the value designed for fully-deployed state $r_{\rm fd}\in\mathbb{R}_+$: 
\begin{equation}
    r_{\rm pd}:=r_{\rm fd}
\end{equation}
and the diameter of its compressed side $r_{\rm pc}\in\mathbb{R}_+$ is set as the minimal value between the deployed diameter, and the addition of compressed diameter and twice gap width:
\begin{equation}
    r_{\rm pc}:={\rm min}\{r_{\rm fc}+2w_{\rm g}, r_{\rm fd}\}
\end{equation}
Using the diameters of the deployed side and the compressed side, a cone shape can be modelled for the partially-deployed stent segment.

Following \cite{zhou2018real_ral}, these circle vertices are accumulated by connecting the neighbouring vertices regularly into triangular faces, resulting in a mathematically modelled stent segment mesh. Fenestrations or scallops are modelled by removing the corresponding vertices and triangular faces. The resolution of height $h$ was set as 0.1mm and that of rotation angle $\theta$ was set as $1^\circ$. A set of five customized markers are sewn on each stent segment. With known pre-operative 3D reference marker positions (3D marker references) and corresponding intra-operative 2D marker positions (2D marker references), the 3D intra-operative pose of marker set which is also the 3D intra-operative pose of the stent segment could be recovered by the RPnP method \cite{zhou2018real_ral}. Details regarding this part will be briefly introduced in Section~\ref{sec:instantiation}.

Unlike the work in \cite{zhoustent} and \cite{zhou2018real_ral} for fully-compressed and fully-deployed stent graft, where 3D marker references are known from computed tomography (CT) scan or stent graft design, 3D marker references for partially-deployed stent segment are unknown due to the unpredictability of the deployment process. 

\subsection{Adapted GCN}
\label{sec:coordinate_prediction}
With known pre-operative 3D fully-deployed marker positions $\textit{\textbf{Y}}_{\rm f}^{\rm l}=(\textit{\textbf{y}}_{{\rm f}1}^{\rm l}~\cdots~\textit{\textbf{y}}_{{\rm f}5}^{\rm l})\in\mathbb{R}^{3\times5}$, an Adapted GCN for regressing pre-operative 3D marker references of partially-deployed stent segment $\textit{\textbf{Y}}_{\rm p}^{\rm l}\in\mathbb{R}^{3\times5}$ is proposed based on \cite{defferrard2016convolutional}. Original GCNs in \cite{defferrard2016convolutional} and \cite{kipf2016semi} were for classification tasks, while in this paper, the coarsening layers are removed and the softmax function at the network end is replaced by linear mapping.

\subsubsection{Data Pre-processing}
To focus the Adapted GCN training on learning the deformation between $\textit{\textbf{Y}}_{\rm f}^{\rm l}$ and $\textit{\textbf{Y}}_{\rm p}^{\rm l}$, in the training data, markers' coordinates for fully-deployed stent segment $\textit{\textbf{Y}}_{\rm f}^{\rm l}$ are standardized in local frame with the transformation: 
\begin{equation}
\textit{\textbf{t}}_{\rm l}^{\rm g}:=\sum_{i=1}^5{({\textit{\textbf{y}}_{\rm f}^{\rm g}}_i)}
\end{equation}
\begin{equation}
\textit{\textbf{R}}_{\rm l}^{\rm g}:= 
\begin{pmatrix}
\textbf{\textit{v}}_1/\|\textbf{\textit{v}}_1\|_2&\textbf{\textit{v}}_2/\|\textbf{\textit{v}}_2\|_2&\textbf{\textit{v}}_3/\|\textbf{\textit{v}}_3\|_2
\end{pmatrix}
\end{equation}
where $\textbf{\textit{v}}_1:={\textit{\textbf{y}}_{\rm f}^{\rm t}}_1$, $\textbf{\textit{v}}_2:=({\textit{\textbf{y}}_{\rm f}^{\rm t}}_1\times{\textit{\textbf{y}}_{\rm f}^{\rm t}}_2)$, $\textbf{\textit{v}}_3:= (\textbf{\textit{v}}_1 \times \textbf{\textit{v}}_2)$ and $\textit{\textbf{Y}}_{\rm f}^{\rm t}:=\textit{\textbf{R}}_{\rm l}^{\rm g} \textit{\textbf{Y}}_{\rm f}^{\rm l}$. $\times$ between two vectors represents the cross product. Then the transformation between global frame and local frame can be represented by:
\begin{equation}
\textit{\textbf{Y}}_{\rm f}^{\rm g}=\textit{\textbf{R}}_{\rm l}^{\rm g}\textit{\textbf{Y}}_{\rm f}^{\rm l}+\textit{\textbf{t}}_{\rm l}^{\rm g}\otimes(\textbf{1})_{1\times5}
\end{equation}
where, $\otimes$ is the kronecker product and $(\textbf{1})_{1\times5}$ is a $1\times5$ matrix consisting of $1$.

Before training the network, the ground truth of markers' coordinates for each partially-deployed stent segment in local frame $\textit{\textbf{Y}}_{\rm p}^{\rm l}$ is obtained by aligning the detected 3D markers' coordinates in global frame $\textit{\textbf{Y}}_{\rm p}^{\rm g}$ to the markers for corresponding fully-deployed stent segment in the local frame $\textit{\textbf{Y}}_{\rm f}^{\rm l}$ via singular value decomposition (SVD): $\textit{\textbf{U}}_{\rm svd}\Sigma{\textit{\textbf{V}}}_{\rm svd}=\textit{\textbf{Y}}_{\rm p}^{\rm g}{\textit{\textbf{Y}}_{\rm f}^{\rm l}}^{\top}$.
The aligned markers' coordinates for each partially-deployed stent segment is thus calculated with mapping $f:(\mathbb{R}^{3\times 5},\mathbb{R}^{3\times 5})\to\mathbb{R}^{3\times 5}$ defined as:
\begin{equation}
\textit{\textbf{Y}}_{\rm p}^{\rm l}=f(\textit{\textbf{Y}}_{\rm p}^{\rm g},\textit{\textbf{Y}}_{\rm f}^{\rm l}):=\textit{\textbf{R}}_{\rm p}^{\rm f}\textit{\textbf{Y}}_{\rm p}^{\rm g}+\textit{\textbf{t}}_{\rm p}^{\rm f}
\end{equation}
where
$
\textit{\textbf{R}}_{\rm p}^{\rm f}:=\textit{\textbf{V}}_{\rm svd}\textit{\textbf{U}}_{\rm svd}^\top
$ and
$
\textit{\textbf{t}}_{\rm p}^{\rm f}:=\sum_{i=1}^5{({\textit{\textbf{y}}_{\rm f}^{\rm l}}_i)}-\textit{\textbf{R}}_{\rm p}^{\rm f}\sum_{i=1}^5{({\textit{\textbf{y}}_{\rm p}^{\rm g}}_i)}
$
are the rotation matrix and translation vector of the transformation.

\subsubsection{Spectral Graph Convolution}
Different from conventional convolutional kernels used in Euclidean space, GCN employs spectral graph convolution on a graph \cite{shuman2012emerging}. The spectral graph Fourier transform and its inverse transform is defined as:
\begin{equation}
\tilde{\textit{\textbf{Y}}}=\mathcal{F}_\mathcal{G}(\textit{\textbf{Y}}):=\textit{\textbf{U}}^\top \textit{\textbf{Y}},\quad \textit{\textbf{Y}}=\mathcal{F}_\mathcal{G}^{-1}(\tilde{\textit{\textbf{Y}}})=\textit{\textbf{U}} \tilde{\textit{\textbf{Y}}}
\end{equation}
where $\mathcal{G}=(\mathcal{V},\mathcal{E},{\textit{\textbf{W}}})$ is an undirected simple graph with $n=5$ nodes, representing the coordinates of five customized markers, $\mathcal{V}$ is a finite set of $|\mathcal{V}|=n$ vertices, $\mathcal{E}\subseteq \mathcal{V}\times\mathcal{V}$ is a set of edges, $\textit{\textbf{W}}\in\mathbb{R}^{n\times n}$ is the weighted adjancy matrix, $\textit{\textbf{Y}}$ is the coordinates' values defined on nodes, the fourier basis $\textit{\textbf{U}}$ is obtained by the eigenvector matrix of graph $\mathcal{G}$'s normalized Laplacian matrix $\textit{\textbf{L}}\in\mathbb{R}^{5\times5}$: $\textit{\textbf{L}}=\textit{\textbf{U}}\Lambda\textit{\textbf{U}}^{-1}$, where $\Lambda={\rm diag}(\lambda_0~\cdots~\lambda_{n-1})\in\mathbb{R}^{n}$ is the eigen values. The normalized Laplacian matrix is defined as:
\begin{equation}
\textit{\textbf{L}}:=\textit{\textbf{D}}^{-0.5}(\textit{\textbf{D}}-\textit{\textbf{W}})\textit{\textbf{D}}^{-0.5}
\end{equation}
where $\textit{\textbf{D}}\in\mathbb{R}^{n\times n}$ is the diagonal degree matrix. As normalized Laplacian matrix is semi-positive definite symmetric matrix, $\textit{\textbf{U}}^\top=\textit{\textbf{U}}^{-1}$. Spectral graph convolution on graph $\mathcal{G}$ could be defined as:
\begin{equation}
\label{eq:convolution}
(\textit{\textbf{g}}_\vartheta*\textit{\textbf{Y}})_\mathcal{G}:=\mathcal{F}^{-1}_\mathcal{G}(\mathcal{F}_\mathcal{G}\big(\textit{\textbf{g}}_\vartheta)\mathcal{F}_\mathcal{G}(Y)\big)=\textit{\textbf{U}}\tilde{\textit{\textbf{g}}}_\vartheta\textit{\textbf{U}}^\top\textit{\textbf{Y}}
\end{equation}
where $\tilde{\textit{\textbf{g}}}_\vartheta$ is defined as the convolutional kernel (also known as filter in \cite{defferrard2016convolutional}) and $\vartheta$ is the trainable parameters. A non-parametric kernel is defined as $\tilde{\textit{\textbf{g}}}_\vartheta(\Lambda)={\rm diag}(\vartheta)$ \cite{bruna2013spectral}, where $\vartheta\in\mathbb{R}^n$. There are also multiple approaches of parametrization for the localized filter, polynomial parametrization was introduced in \cite{defferrard2016convolutional}: $\tilde{\textit{\textbf{g}}}_\vartheta(\Lambda)=\sum_{k=0}^{K-1}{\vartheta_k (\Lambda)^k}$, where ${(\Lambda)^k}$ is the $k$ power of $\Lambda$. Because $\textit{\textbf{U}}^\top=\textit{\textbf{U}}^{-1}$ and this polynomial parametric kernel converts (\ref{eq:convolution}) into:
\begin{equation}
(\textit{\textbf{g}}_\vartheta*\textit{\textbf{Y}})_\mathcal{G}=\tilde{g}_\vartheta(\textit{\textbf{L}})\textit{\textbf{Y}}=\sum_{k=0}^{K-1}{\vartheta_k (\textbf{\textit{L}})^k}\textit{\textbf{Y}}
\end{equation}
where $K\in\mathbb{Z}_+$ represents the kernel size and $\vartheta\in\mathbb{R}^K$ implies the learning complexity to be reduced to $\mathcal{O}(K)$, compared with $\mathcal{O}(n)$ for non-parametric kernel.

Furthermore, recursive formulation for parametric kernel was introduced in \cite{defferrard2016convolutional} to reduce the computational time. The kernel is approximated by Chebyshev polynomials:
\begin{equation}
\tilde{\textit{\textbf{g}}}_\vartheta(\Lambda)=\sum_{k=0}^{K-1}{\vartheta_k T_k(\Lambda')}
\end{equation}
where $\Lambda'=2\Lambda/\lambda_{\rm max}-\textit{\textbf{I}}_n$, $\textit{\textbf{I}}_n$ is an unity matrix with size $n\times n$. $T_k(\Lambda')=2\Lambda'T_{k-1}(\Lambda')-T_{k-2}(\Lambda')$ is the recursive Chebyshev polynomials with $T_0(\Lambda')=1$ and $T_1(\Lambda')=\Lambda'$. This kernel is used in this paper and details can be found in \cite{hammond2011wavelets}.

\subsubsection{Network Architecture}
The number of five customized markers is shown in Figure~\ref{fig:intro}(d). An undirected simple graph $\mathcal{G}=(\mathcal{V},\mathcal{E},\textit{\textbf{W}})$ with five nodes is constructed to represent the five markers' coordinates with weight adjacency matrix set referring to the distance scale:
\begin{equation}
\textit{\textbf{W}}=\begin{pmatrix}
0&e^{-(5/4)^2}&0&0&e^{-(5/8)^2}\\
e^{-(5/4)^2}&0&e^{-(5/4)^2}&0&0\\
0&e^{-(5/4)^2}&0&e^{-(5/4)^2}&0\\
0&0&e^{-(5/4)^2}&0&e^{-(5/4)^2}\\
e^{-(5/8)^2}&0&0&e^{-(5/4)^2}&0\\
\end{pmatrix}    
\end{equation}

\begin{figure}[ht]
     \centering
      \framebox{\parbox{2.9in}{\includegraphics[width = 2.9in]{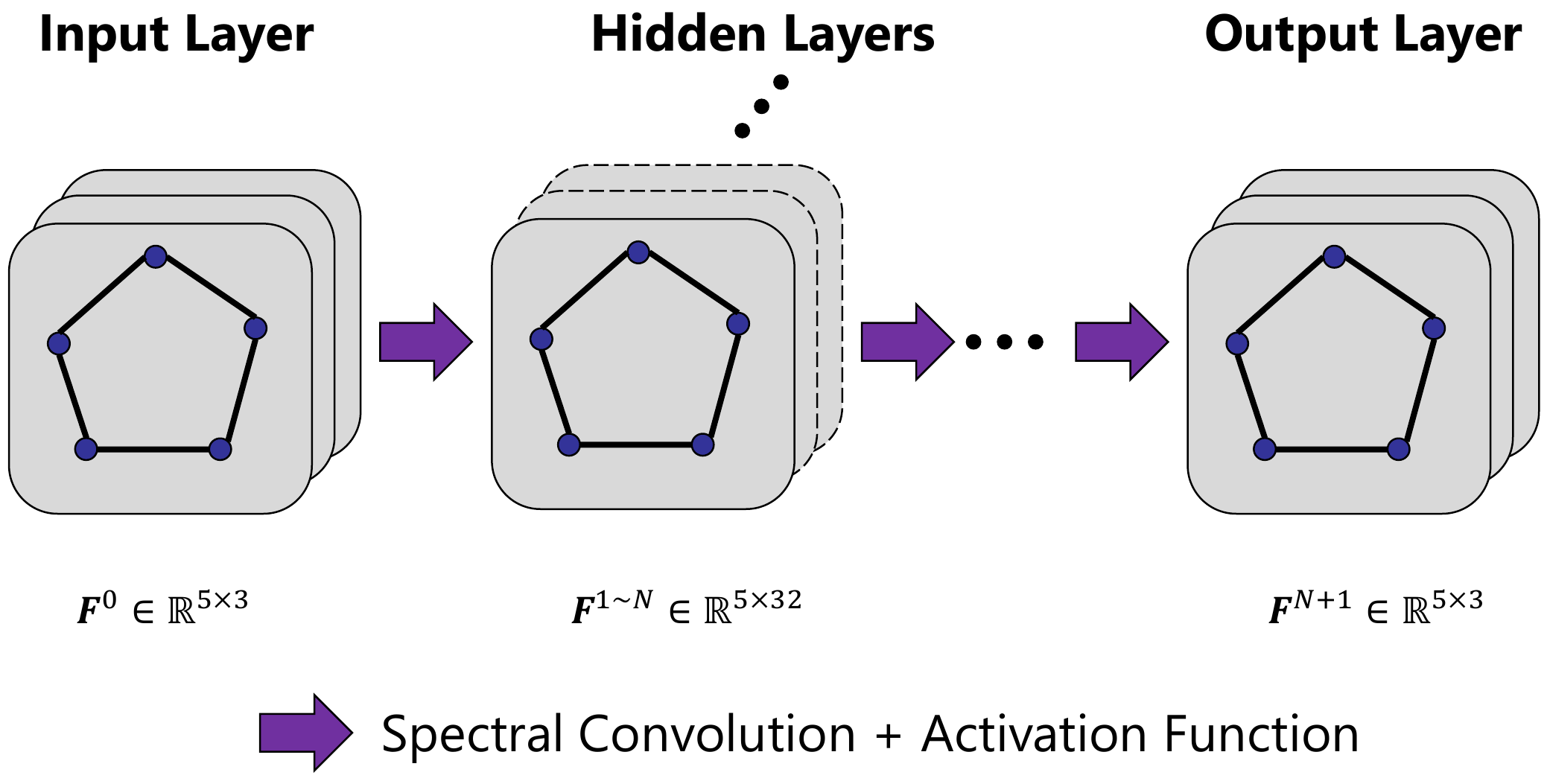}}}
      \caption{Network architecture of the proposed Adapted GCN.}
      \label{fig:network}
\end{figure}

The network architecture is shown in Figure~\ref{fig:network}, where the input is ${\textit{\textbf{Y}}}_{\rm f}^{\rm l}+\epsilon$ and the output is $\hat{\textit{\textbf{Y}}}_{\rm p}^{\rm l}$, where $\epsilon\sim\mathcal{N}(0,0.1)$ is Gaussian noise. The mathematical expression for each two neighbouring layers can be written as:
\begin{equation}
    \textit{\textbf{F}}^i=\sigma_i\big((\textit{\textbf{g}}_\vartheta*\textit{\textbf{F}}^{i-1})_\mathcal{G}\big)
\end{equation}
where $i\in[0,N+1]\cap\mathbb{Z}$, $\textit{\textbf{F}}^0$ is the input graph, $\textit{\textbf{F}}^{N+1}$ is the output graph, $\textit{\textbf{F}}^{1\sim N}$ are hidden layers, $N$ is the hidden layer number and $\sigma_i(\cdot)$ is the activation function for the $i^{\rm th}$ layer.

Eight hidden layers are used for the experiments, 32 channels are set in each hidden layer. Leaky ReLU is used as the activation function for non-linear mapping with 0.1 leaky rate for the input and the hidden layers. No non-linear activation function is used in the output layer. Chebyshev polynomial parametric kernel is used with an kernel size of 2 for each spectral convolutional layer.

\subsubsection{Loss Function and Optimization}
The root mean square error between the ground truth and the output coordinates is calculated as the loss function, with a regularization term of L2 norm of the weight matrix:
\begin{equation}
    \mathcal{L}=\|\hat{\textit{\textbf{Y}}}_{\rm p}^{\rm l}-\textit{\textbf{Y}}_{\rm p}^{\rm l}\|_2+\alpha\|\vartheta\|_2
\end{equation}
Adam and Momentum Stochastic Gradient Descend (SGD) were compared for training the network. The optimization through Adam was hard to converge and hence Momentum SGD was used as the optimizer. The learning rate was set as 0.0001 and the learning momentum was set as 0.9. The L2 norm weight $\alpha$ was set as $5\times 10^{-4}$ and the batch size was set as 10.

As the RPnP method is only related to 3D reference marker shapes while is free to global 3D reference marker positions, the predicted 3D marker references $\hat{\textit{\textbf{Y}}}_{\rm p}^{\rm l}$ are also aligned to the local markers' coordinates of fully-deployed stent segment ${\textit{\textbf{Y}}}_{\rm f}^{\rm l}$ as $f(\hat{\textit{\textbf{Y}}}_{\rm p}^{\rm l},{\textit{\textbf{Y}}}_{\rm f}^{\rm l})$ for the transformation estimation of partially-deployed stent segments. 

\subsection{3D Shape Instantiation}
\label{sec:instantiation}
With the predicted pre-operative 3D marker references from the Adapted GCN in Section~\ref{sec:coordinate_prediction} and manually labelled corresponding intra-operative 2D marker positions/references, following \cite{zhou2018real_ral}, the RPnP method \cite{li2012robust} is used to instantiate the 3D pose of intra-operative marker set including the rotation matrix $\hat{\textit{\textbf{R}}}_{\rm l}^{\rm g}\in\mathbb{R}^{3\times3}$ and translation vector $\hat{\textit{\textbf{t}}}_{\rm l}^{\rm g}\in\mathbb{R}^{3}$:
\begin{equation}
    \hat{\textit{\textbf{Y}}}_{\rm p}^{\rm g}=\hat{\textit{\textbf{R}}}_{\rm l}^{\rm g}f(\hat{\textit{\textbf{Y}}}_{\rm p}^{\rm l},{\textit{\textbf{Y}}}_{\rm f}^{\rm l})+\hat{\textit{\textbf{t}}}_{\rm l}^{\rm g}\otimes(\textbf{1})_{1\times5}
\end{equation}
where $\hat{\textit{\textbf{Y}}}_{\rm p}^{\rm g}$ is the instantiated intra-operative 3D marker positions for partially-deployed stent segment. As markers are sewn on the stent segment, $\hat{\textit{\textbf{R}}}_{\rm l}^{\rm g}$ and $\hat{\textit{\textbf{t}}}_{\rm l}^{\rm g}$ are also the rotation matrix and translation vector for the partially-deployed stent segment. After moving the mathematically modelled stent segment mesh in Section~\ref{sec:stent_model} to the same local coordinate frame, $\hat{\textit{\textbf{R}}}_{\rm l}^{\rm g}$ and $\hat{\textit{\textbf{t}}}_{\rm l}^{\rm g}$ are applied for the stent segment transformation. After central point based correction, 3D shape instantiation of partially-deployed stent segment is achieved. More details could be found in \cite{zhou2018real_ral}.


\subsection{Experiment and Validation}
\label{sec:experiment}

\subsubsection{Marker Design}
Customized stent graft markers with five different shapes were designed based on commercially-used gold markers and were manufactured on a Mlab Cusing R machine (ConceptLaser, Lichtenfels, Germany) from SS316L stainless steel powder, as shown in Figure~\ref{fig:intro}(d) with their own numbers. The sizes are around 1$\sim$3 mm, similar to the commercial ones. Those five markers were sewn on each stent segment at five non-planar places.

\begin{figure}[ht]
     \centering
      \framebox{\parbox{3.3in}{\includegraphics[width = 3.3in]{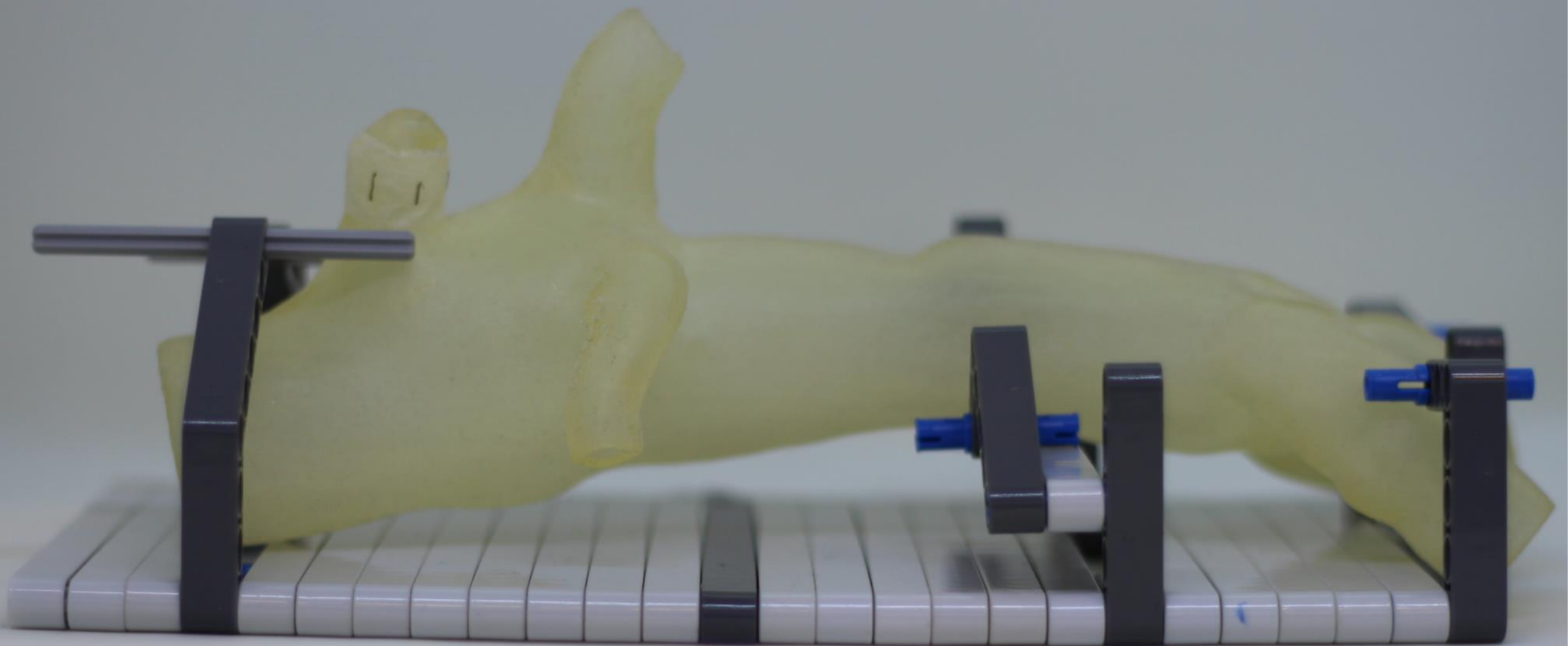}}}
      \caption{Illustration of the experimental setup with fixing an AAA phantom under the CT scan.}
      \label{fig:setup}
\end{figure}

\subsubsection{Simulation of Surgery}
Three stent grafts were used in the experiments, including a iliac stent graft (Cook Medical, IN, USA) with five stent segments, 10$\sim$19mm diameters and total $90$mm height, a fenestrated stent graft (Cook Medical) with six stent segments, 22$\sim$30mm diameters and total 117mm height, and a thoracic stent graft (Medtronic, MN, USA) with 10 stent segments, 30mm diameter and total 179mm height. Five AAA phantoms were modelled from CT data scanned from patients and were printed on a Stratasys Object 3D printer (MN, USA) with VeroClear and TangoBlack colours. To simulate the practical situation in FEVAR where the fenestrated stent graft is customized to similar diameters to that of the AAAs, two suitable AAA target positions where their diameters are similar to that of the corresponding experiment stent graft were selected for each experiment stent graft, resulting in 6 experiments in total. The selected AAA phantom was fixed as shown in Figure~\ref{fig:setup}. In each experiment, a stent graft was compressed into a Captivia delivery catheter (Medtronic) with 8mm diameter, inserted into the selected phantom and deployed subsequently segment-by-segment from the proximal end to the distal end at the target AAA position. 

\subsubsection{Data Collection}
A 3D CT scan and a 2D fluoroscopic image at the frontal plane were scanned for each partially-deployed stent graft using a GE Innova 4100 (GE Healthcare, Bucks, UK) system. The stent segments at the distal end and with odd indexes in the thoracic stent graft experiment were ignored to keep data balance. Thus, there are eight partially-deployed stent segments scanned by CT and flurorscopy in two different AAA phantoms for the iliac stent graft (segment number 1-4 and 5-8), 10 for the fenestrated stent graft (stent segment number 9-13 and 14-18), and eight for the thoracic stent graft (stent segment number 19-22 and 23-26). In addition, three CT scans were acquired for the three experiment stent grafts at fully-deployed state to supply 3D fully-deployed marker positions - $\textit{\textbf{Y}}_{\rm f}^{\rm l}$. In practical applications, this information can be obtained from stent graft designing.


\subsubsection{Marker Position Extraction}
\label{sec:marker_detection}
Although Equally-weighted Focal U-Net was proposed to potentially achieve automatic 2D marker segmentation and classification from intra-operative 2D fluoroscopic images. In this paper, the stent graft is in partially-deployed state which is different from the training data in \cite{zhou2018towards_iros} where the stent graft was in fully-deployed state. The segmentation and classification results of applying the trained model in \cite{zhou2018towards_iros} onto the fluoroscopic images in this paper is not accurate and unsatisfied. Hence the intra-operative 2D marker positions or references $\textit{\textbf{X}}^{\rm g}=(\textit{\textbf{x}}_1^{\rm g}~\cdots~\textit{\textbf{x}}_5^{\rm g})\in\mathbb{R}^{2\times5}$ were extracted manually via $Matlab^{\textregistered}$. 

The shapes of 3D stents and 3D customized markers were segmented from CT scans via ITK-SNAP and the 3D central coordinates of customized markers $\textit{\textbf{Y}}^{\rm g}=(\textit{\textbf{y}}_1^{\rm g}~\cdots~\textit{\textbf{y}}_5^{\rm g})\in\mathbb{R}^{3\times5}$ were extracted using Meshlab.

\subsubsection{Data Augmentation}
Before training the Adapted GCN with the 3D marker positions of fully-deployed and partially-deployed stent segments, these coordinates were rotated and scaled to enlarge the training dataset. The rotations about three axises range from $-30^\circ$ to $30^\circ$ with the resolution of $3^\circ$. The scale ratios range from $0.2$ to $11.39$ with the geometric proportion of $1.5$.

\subsubsection{Criteria and Evaluation}
To evaluate the 3D marker references predicted by the proposed Adapted GCN, the aligned 3D marker reference prediction $\textit{\textbf{Y}}_{\rm p}^{\rm l}$ were compared to the ground truth of the aligned partially-deployed stent segment's marker positions $f(\hat{\textit{\textbf{Y}}}_{\rm p}^{\rm l},{\textit{\textbf{Y}}}_{\rm f}^{\rm l})$ via their mean distance error, ${\rm MDE}\big(\textit{\textbf{Y}}_{\rm p}^{\rm l},f(\hat{\textit{\textbf{Y}}}_{\rm p}^{\rm l},{\textit{\textbf{Y}}}_{\rm f}^{\rm l})\big)$, which is calculated as:
\begin{equation}
    {\rm MDE}(\textit{\textbf{Y}}^1,\textit{\textbf{Y}}^2)=\frac{1}{n}\sum_{i=1}^{n}{\big\|\textit{\textbf{y}}_i^1-\textit{\textbf{y}}_i^2\big\|_2}
\end{equation}
where $\textit{\textbf{Y}}^{1}$ and $\textit{\textbf{Y}}^{2}$ can be two matrices of 3D or 2D marker coordinates with the same dimension number and the same point number.

To evaluate marker instantiation, the registered global markers' coordinates for each partially-deployed stent segment $\hat{{\textit{\textbf{Y}}}}_{\rm p}^{\rm g}$ are compared with the ground truth ${{\textit{\textbf{Y}}}}_{\rm p}^{\rm g}$ via ${\rm MDE}\big({{\textit{\textbf{Y}}}}_{\rm p}^{\rm g},\hat{{\textit{\textbf{Y}}}}_{\rm p}^{\rm g}\big)$ in 3D and the reprojected distance error ${\rm MDE}\big({{\textit{\textbf{X}}}}_{\rm p}^{\rm g},\hat{{\textit{\textbf{X}}}}_{\rm p}^{\rm g}\big)$ in 2D, where $\hat{{\textit{\textbf{X}}}}_{\rm p}^{\rm g}$ is the projected 2D coordinate from the estimated 3D global coordinate $\hat{{\textit{\textbf{Y}}}}_{\rm p}^{\rm g}$, calculated by $\hat{{\textit{\textbf{X}}}}_{\rm p}^{\rm g}=g(\hat{{\textit{\textbf{Y}}}}_{\rm p}^{\rm g})$ with mapping $g:\mathbb{R}^{3\times n}\to\mathbb{R}^{2\times n}$:
\begin{equation}
\label{eq:projection_points}
g(\textit{\textbf{Y}})=\begin{pmatrix}
    \textit{\textbf{p}}_1^\top\textit{\textbf{Y}}^{\rm h}\oslash\textit{\textbf{p}}_3^\top\textit{\textbf{Y}}^{\rm h}	\\	\textit{\textbf{p}}_2^\top\textit{\textbf{Y}}^{\rm h}\oslash\textit{\textbf{p}}_3^\top\textit{\textbf{Y}}^{\rm h}
	                   \end{pmatrix}
\end{equation}
where $\textit{\textbf{P}}=\begin{pmatrix} \textit{\textbf{p}}_1	&	\textit{\textbf{p}}_2	&	\textit{\textbf{p}}_3 \end{pmatrix}^\top\in\mathbb{R}^{3\times4}$ is the projection matrix, $\oslash$ is Hadamard division, and $\textit{\textbf{Y}}^{\rm h}=(\textit{\textbf{y}}_1^{\rm h}~\cdots~\textit{\textbf{y}}_{\textit{n}^3}^{\rm h})=(\textit{\textbf{Y}}^\top~(\textbf{1})_{\textit{n}^3\times1})^\top\in\mathbb{R}^{{4}\times{\textit{n}^3}}$ is the homogeneous vector form of the 3D coordinates.

To evaluate 3D shape instantiation for each partially-deployed stent segment, the distance between the instantiated partially-deployed stent segment mesh and the corresponding ground truth was measured using Matlab function $point2trimesh$ \cite{point2trimesh}. Marker angle was estimated by the angle of the nearest vertex on the constructed stent segment. Mean absolute angle difference between the predicted markers and the ground truth was used to measure the angle error.

\subsubsection{Cross Validation}
Three-fold cross validations were performed along the division of stent graft. For example, for testing stent segments on iliac stent grafts, the data from the fenestrated and thoracic stent graft were used for the training.

\section{Results}
\label{sec:result}
In this section, the experimental results for the validation of the proposed method was illustrated including the 3D distance errors in the marker prediction, the 2D re-projected and 3D distance error in the marker instantiation, as well as the angular and the mesh error in the stent segment shape instantiation, 

\subsection{Prediction of 3D Marker References}
\label{sec:result_GCN}

\begin{figure}[ht]
     \centering
      \framebox{\parbox{3.3in}{\includegraphics[width = 3.3in]{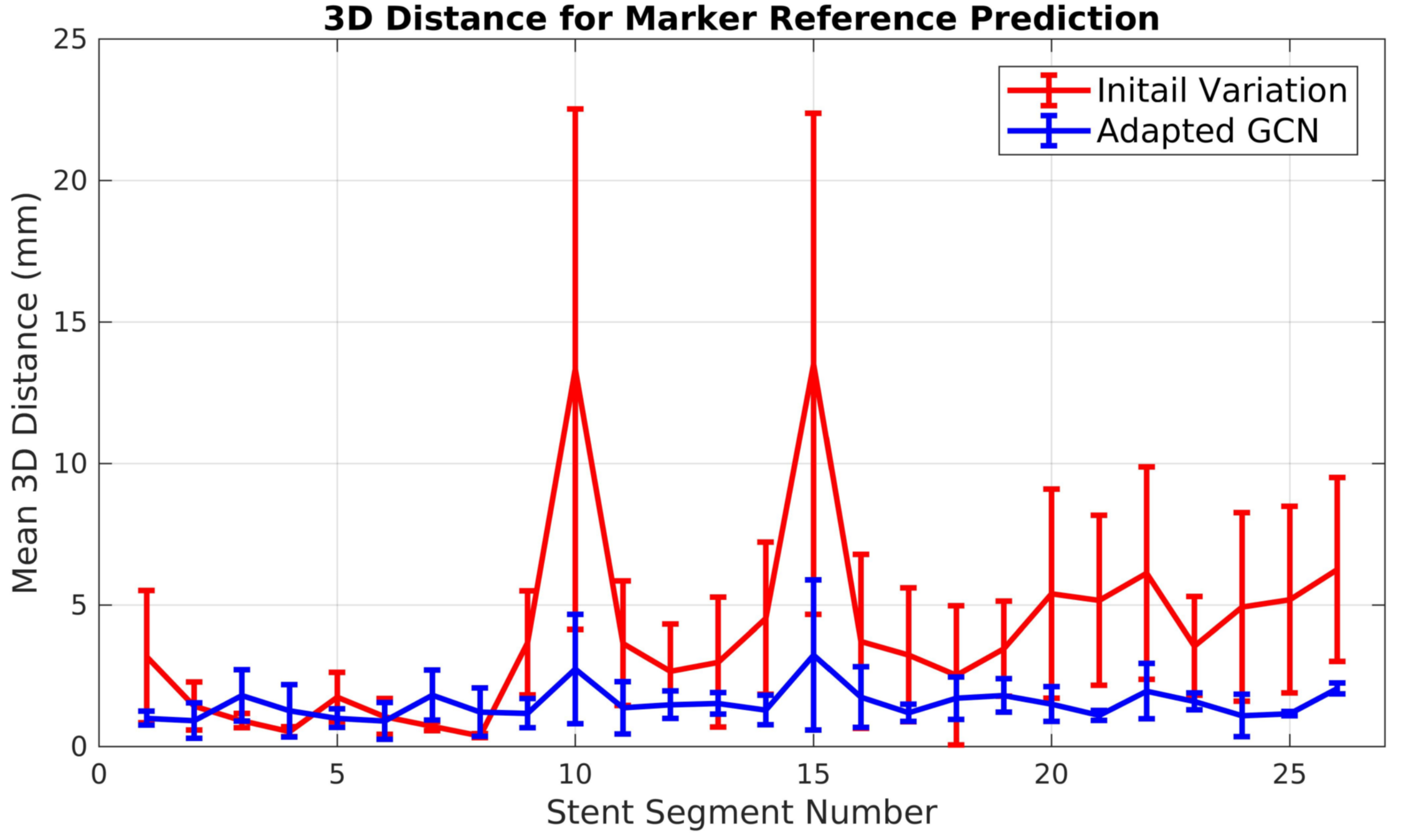}}}
      \caption{Mean$\pm$std 3D distance of the initial variation and mean$\pm$std 3D distance error of 3D marker reference prediction with the proposed Adapted GCN.}
      \label{fig:deform_error}
\end{figure}

The mean 3D distance between the prediction of 3D marker references and the ground truth, called Adapted GCN, and the initial mean 3D distance between the 3D fully-deployed markers and the ground truth, named initial variation, for the 26 partially-deployed stent segments are shown in Figure~\ref{fig:deform_error}. We can see that the mean 3D distance achieved by the Adapted GCN is significantly lower than the initial variation, especially for the fenestrated and thoracic stent graft (stent segment number 9$\sim$26), proving the efficiency of the proposed Adapted GCN on 3D marker reference prediction. The mean 3D distances achieved by the Adapted GCN for the iliac stent graft (stent segment number 1$\sim$8) are comparable to the initial variations. Because the diameter of the iliac stent graft is very close to that of the deployment catheter (due to limited experimental resources, we only got one available deployment catheter), and there is not much difference between the fully-deployed and partially-deployed state of the iliac stent graft.

\subsection{3D Marker Instantiation}

\begin{figure}[th]
     \centering
      \framebox{\parbox{3.3in}{\includegraphics[width = 3.3in]{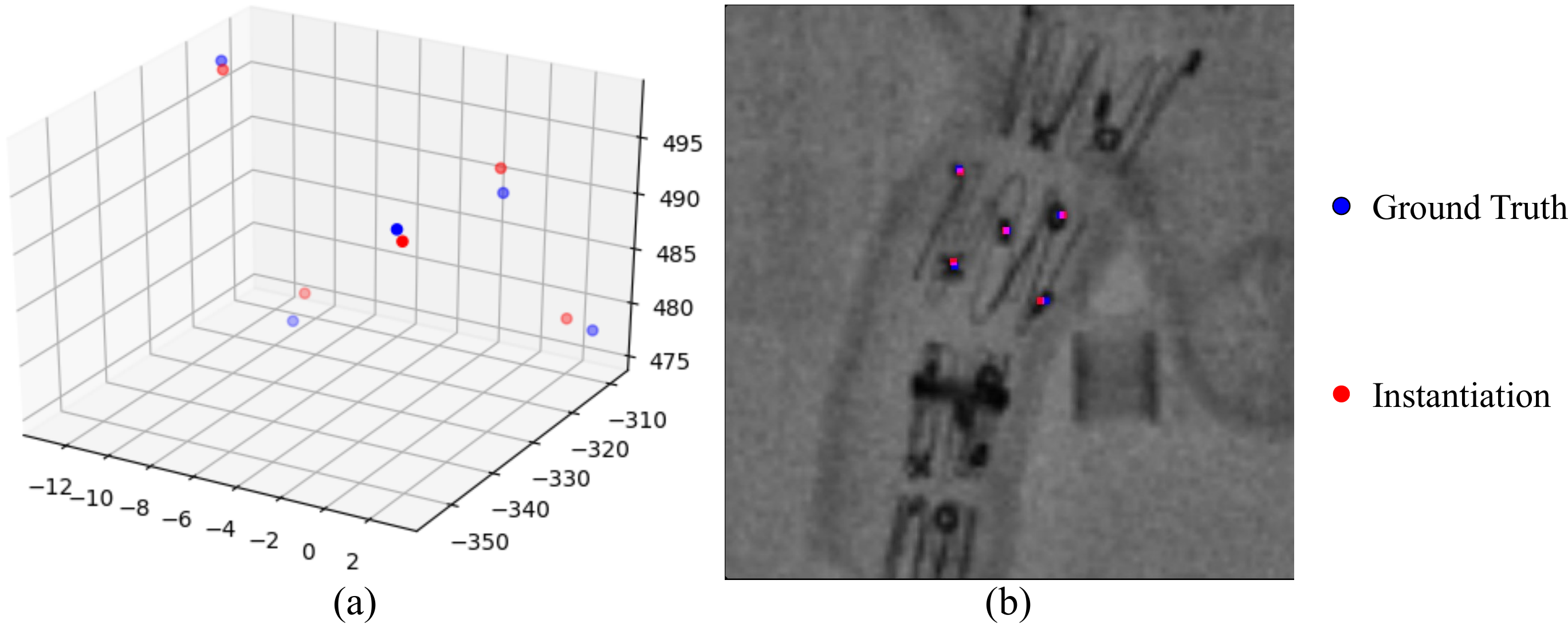}}}
      \caption{Comparison of instantiated intra-operative 3D marker positions and the 3D ground truth (a), and comparison of 2D projections of instantiated 3D markers and the 2D ground truth (b).}
      \label{fig:pose_example}
\end{figure}

The predicted 3D marker references and the manually detected 2D marker references for partially-deployed stent segment are imported into the RPnP instantiation framework \cite{zhou2018real_ral} to recover the intra-operative 3D marker positions. The instantiated intra-operative 3D marker positions and their 2D projections are compared to the corresponding ground truth, with results shown in Figure~\ref{fig:pose_example}. We can see that the instantiated marker positions are very close to the ground truth in both 3D and 2D.

\begin{figure}[th]
     \centering
      \framebox{\parbox{3.3in}{\includegraphics[width = 3.3in]{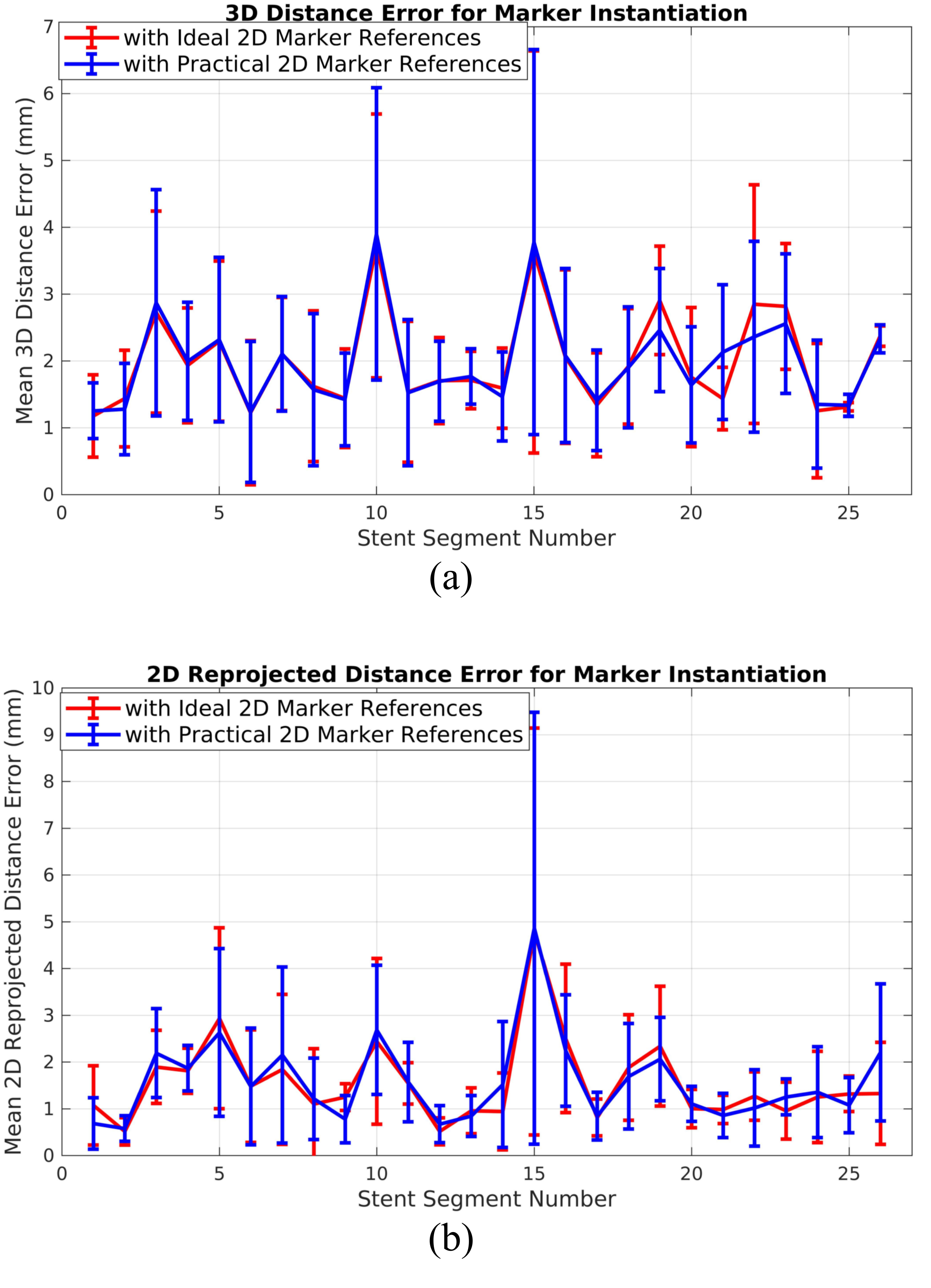}}}
      \caption{Mean$\pm$std 3D (a) and 2D projected (b) distance errors of the instantiated intra-operative marker positions with the ideal (red) and practical (blue) 2D marker references as the input 2D marker reference.}
      \label{fig:pose_error}
\end{figure}

Due to the imaging error caused by the fluoroscopic system, 0.5$\sim$0.8mm deviation exists between the manually detected 2D marker references, named practical 2D marker references, and the projected 2D marker references from the ground truth 3D marker references, named ideal 2D marker references. Both of these two 2D marker references are used with the predicted 3D marker references to instantiate the intra-operative 3D marker positions. The 3D and 2D re-projected distance errors for the 26 partially-deployed stent segments are shown in Figure~\ref{fig:pose_error}. We can see that an average 2D distance error of 1.58mm and an average 3D distance error of 1.98mm are achieved respectively. The small accuracy gap in the Figure~\ref{fig:pose_error} between using practical and ideal 2D marker references indicates that the robustness of the instantiation framework to the imaging error introduced by the fluoroscopic system.

\begin{figure}[th]
     \centering
      \framebox{\parbox{3.3in}{\includegraphics[width = 3.3in]{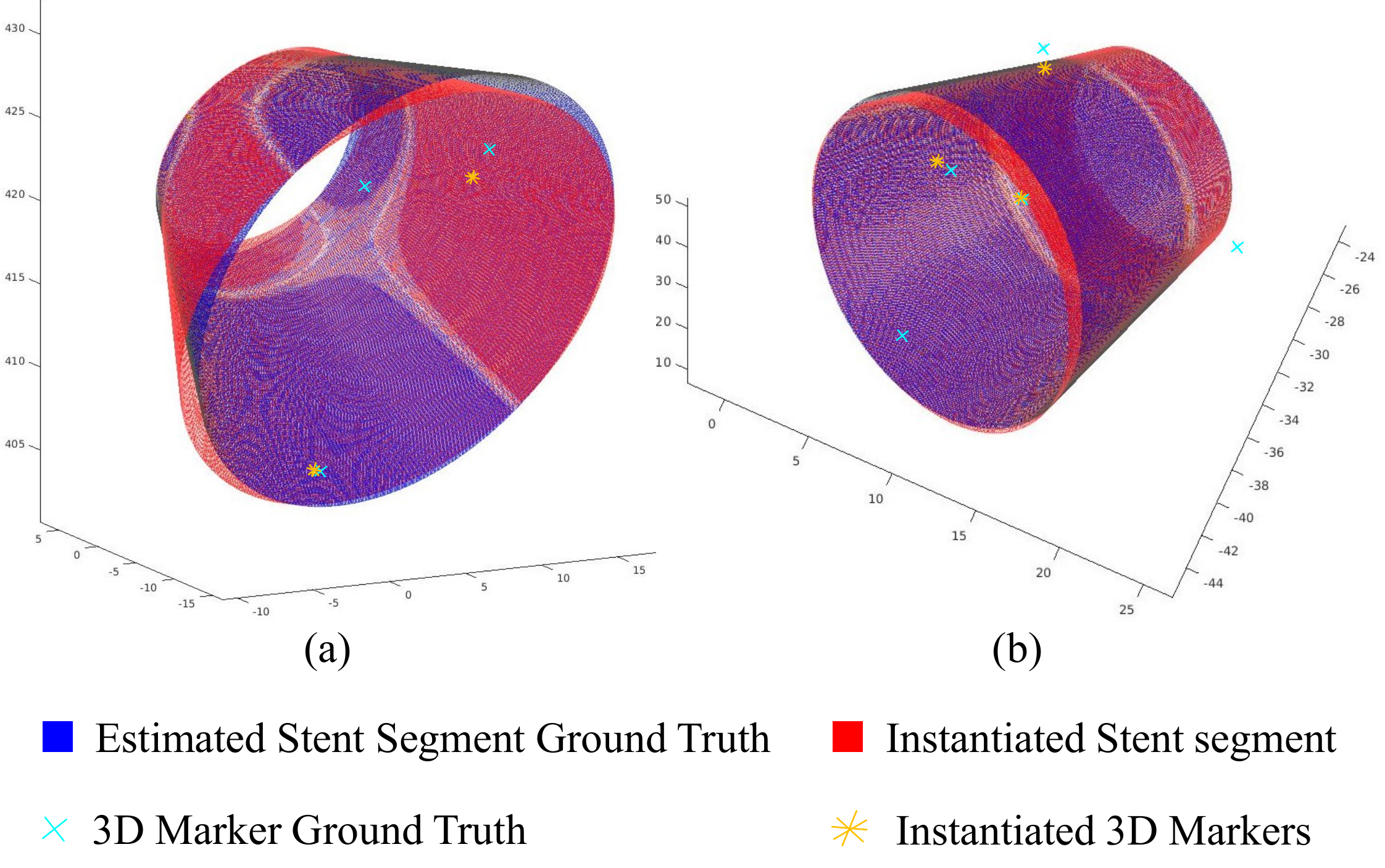}}}
      \caption{Two comparison examples of instantiated meshes of partially-deployed stent segment and 3D makers from predicted 3D and practical 2D marker references, compared with the estimated stent segment ground truth and the 3D marker ground truth.}
      \label{fig:mesh_example}
\end{figure}

\begin{figure}[th]
     \centering
      \framebox{\parbox{3.3in}{\includegraphics[width = 3.3in]{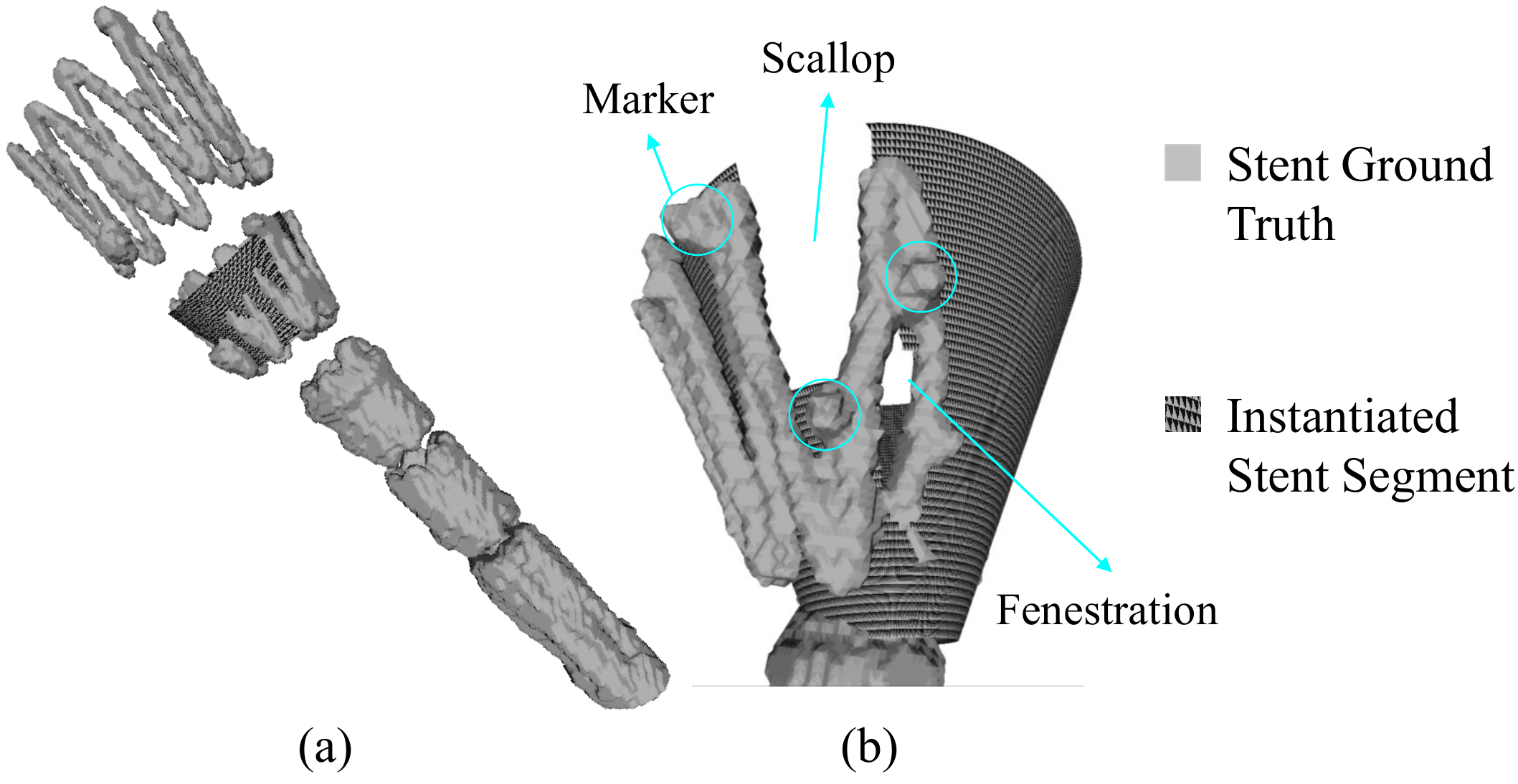}}}
      \caption{Two comparison examples of instantiated meshes of partially-deployed stent segment from predicted 3D and practical 2D marker references, compared with the corresponding stent ground truth segmented from CT scan.}
      \label{fig:stent_example}
\end{figure}

\subsection{3D Shape Instantiation of Partially-deployed Stent Segment}
As graft could not be imaged via CT, the ground truth of partially-deployed stent segment was estimated by registering the mathematical model in Section~\ref{sec:stent_model} onto the ground truth 3D marker references. Two comparison examples of the instantiated partially-deployed stent segment and the estimated ground truth are shown in Figure~\ref{fig:mesh_example}. Two comparison examples of the instantiated partially-deployed stent segment and the real ground truth represented by the CT stent scan are shown in Figure~\ref{fig:stent_example}. We can see that the reasonable 3D shape instantiation is achieved.

\begin{figure}[th!]
     \centering
      \framebox{\parbox{3.3in}{\includegraphics[width = 3.3in]{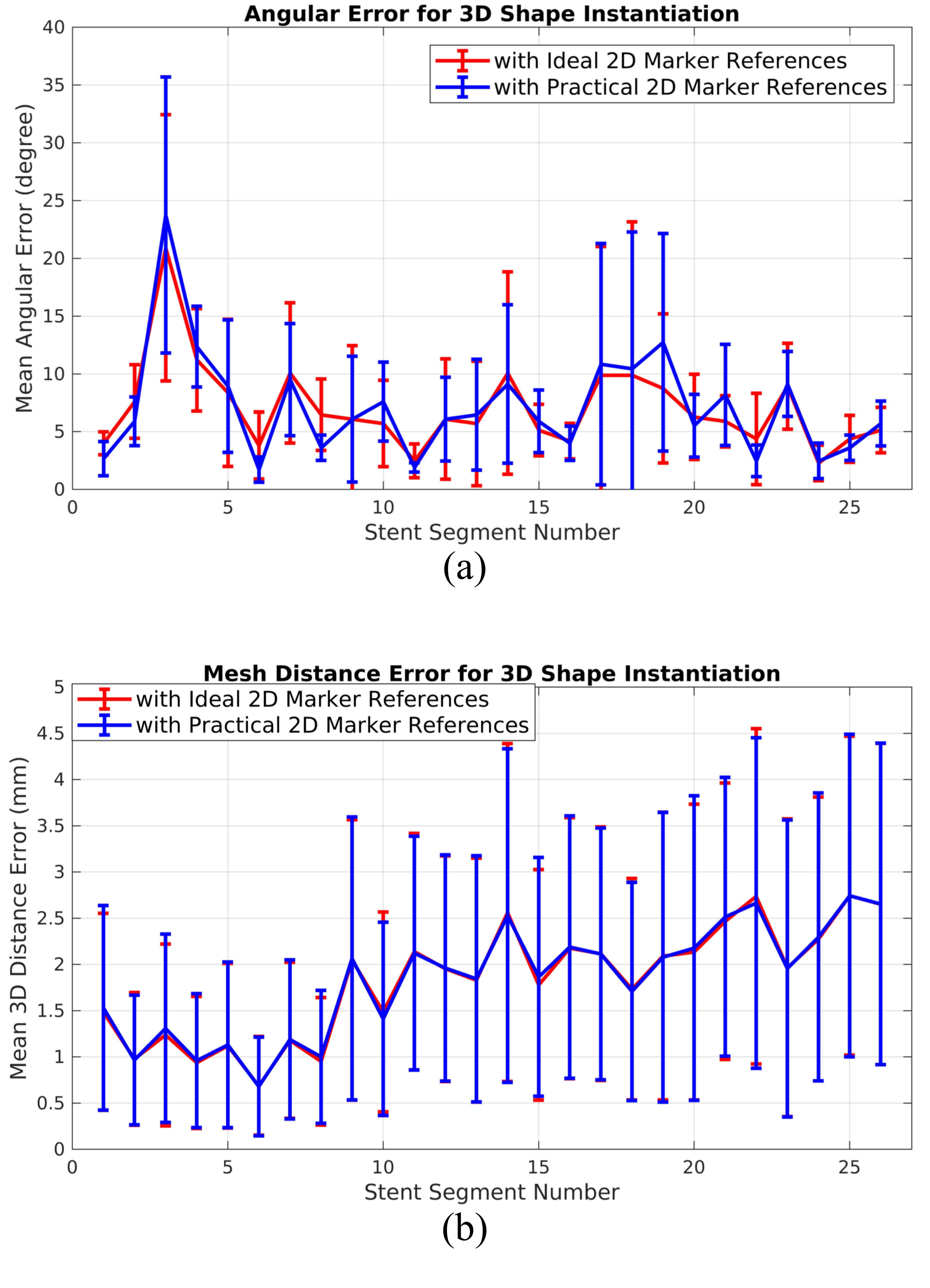}}}
      \caption{Mean$\pm$std angular and 3D mesh distance error of instantiated meshes of partially-deployed stent segment with ideal and practical 2D marker references as the input 2D marker reference.}
      \label{fig:mesh_error}
\end{figure}

The mean angular error between the instantiated intra-operative 3D markers and the ground truth is shown in Figure~\ref{fig:mesh_error}(a). An average angular error of $7^\circ$ is achieved which is larger than the average angular error of $4^\circ$ in \cite{zhou2018real_ral}. This is reasonable, as 3D marker references in this paper are unknown and are predicted by training an Adapted GCN. The mean angular error for iliac stent graft (stent segment number 1$\sim$8) is larger than that for the fenestrated and thoracic stent graft (stent segment number 9$\sim$26) due to the same reason stated in Section~\ref{sec:result_GCN}. The mean distance error between the instantiated stent segment mesh and the ground truth is shown in Figure~\ref{fig:mesh_error}(b). An average distance error of 1$\sim$3mm is achieved which is comparable to the average distance error of 1$\sim$3mm in \cite{zhou2018real_ral}. The iliac stent graft (stent segment number 1$\sim$8) experiences lower mean distance error than the fenestrated and thoracic stent graft (stent segment number 9$\sim$26), as its size is smaller.

\begin{table*}[th]
	\centering
	\caption{The overall performance of marker reference prediction, marker instantiation and 3D shape instantiation on the six experiments, via mean 3D distance error (3D dist.), mean 2D projected distance error (2D dist.), angular error (Ang. error) and mesh distance error (Mesh dist.).}
    \label{tab:results}
    \begin{tabular}{ccccccccc}
        \hline
        \multicolumn{3}{c}{Stent graft}      &iliac &iliac  &fenestrated  &fenestrated  &thoracic  &thoracic  \\
        \multicolumn{3}{c}{Stent segment number}    &1-4  &5-8  &9-13  &14-18  &19-22  &23-26  \\
        \hline
        \multirow{2}{*}{Marker references}  &\multirow{2}{*}{3D dist. (mm)}&Initial Variation    &1.5152   &0.9772   &5.2585   &5.5062   &5.0397   &4.9839   \\ \cline{3-9}
                                                &                       &Adapted GCN   &1.2490   &1.2374  &1.6595   &1.8378   &1.5935   &1.4778   \\
        \hline
        \multirow{4}{*}{Marker instantiation}    &\multirow{2}{*}{2D dist. (mm)}  &Ideal 2D Marker Reference  &1.3247   &1.8414   &1.3421   &2.1870   &1.3989   &1.2145  \\ \cline{3-9}
                                            &                           &Practical 2D Marker Reference    &1.3300   &1.8671   &1.3101   &2.2328   &1.2607   &1.4742   \\ \cline{2-9}
                                            &\multirow{2}{*}{3D dist. (mm)}  &Ideal 2D Marker Reference   &1.8196   &1.8120   &2.0238   &2.1100   &2.2377&1.9398   \\ \cline{3-9}
                                            &                           &Practical 2D Marker Reference    &1.8505   &1.8085   &2.0629   &2.1285   &2.1495   &1.8948   \\
        \hline
        \multirow{4}{*}{Shape Instantiation}&\multirow{2}{*}{Ang. error ($^\circ$)}   &Ideal 2D Marker Reference   &10.9250   &7.1725   &5.2060   &7.8280   &6.3175&5.1775   \\ \cline{3-9}
                                            &                           &Practical 2D Marker Reference    &11.1625   &5.9375   &5.6240   &8.0560   &7.2200   &5.2250   \\ \cline{2-9}
                                            &\multirow{2}{*}{Mesh dist. (mm)}  &Ideal 2D Marker Reference   &1.1530   &0.9841   &1.8910   &2.0721   &2.3562    &2.4084   \\ \cline{3-9}
                                            &                           &Practical 2D Marker Reference    &1.1688   &0.9992   &1.8803   &2.0800   &2.3579   &2.4122   \\
        \hline
    \end{tabular}
\end{table*}
 
Furthermore, the 3D distance error for 3D marker reference prediction, the 2D projected and 3D distance error for intra-operative 3D marker instantiation, the angular and distance error for 3D shape instantiation for partially-deployed stent segment for each experiment are shown with details in the Tab.~\ref{tab:results}.

For instantiating each stent segment on a computer with a CPU of $Intel^{\textregistered}$ Core(TM) i7-4790 @3.60GHz$\times$8, the computational time is around 7ms using Matlab. The 3D marker reference prediction in Tensorflow on a $Nvidia^{\textregistered}$ Titan Xp GPU costs around 0.8ms for each stent segment. The training of Adapted GCN takes approximately 5 hours. The implemented code was written based on the work of \cite{kipf2016semi}.

\section{Discussion}
\label{sec:discussion}
In this paper, a 3D shape instantiation approach based on a previously deployed framework \cite{zhou2018real_ral} is proposed for partially-deployed stent segment from a single intra-operative 2D fluoroscopic image. It is validated on three commonly used stent grafts with five different AAA phantoms. The mean distance errors of instantiated stent segments are around 1$\sim$3mm and the mean angular errors of instantiated markers are around $5^\circ \sim 11^\circ$.

Without knowing pre-operative 3D marker references, the Adapted GCN is introduced into the previous shape instantiation framework \cite{zhou2018real_ral} and achieves reasonable 3D marker reference prediction (an average 3D distance error of 1.5mm for the fenestrated and thoracic stent graft) from 3D fully-deployed markers. However, the 3D marker reference prediction for the iliac stent graft is insufficient. The diameter of deployment catheter used in the experiments is almost the same as that of the iliac stent graft, resulting in the partially-deployed 3D marker set shape is almost the same as the fully-deployed one. In the cross validation for the iliac stent graft, the Adapted GCN was trained on the fenestrated and thoracic stent graft data for learning partially-deployed deformation. The trained model would not be suitable for predicting 3D marker references for the iliac stent graft which did not experience obvious partially-deployed deformation.

In the training of the Adapted GCN, batch normalization and dropout were also explored, but these two methods decreased the accuracy. One potential reason for the batch normalization's performance is the network for regression tasks is sensitive to the scale of feature value and thus the usage of batch normalization in this task should be different. Future work is essential to confirm the feasebility of batch normalization and dropout in the proposed Adapted GCN.

The errors of 3D marker or shape instantiation with using ideal and practical 2D marker references are very similar in Figure~\ref{fig:pose_error} and Figure~\ref{fig:mesh_error}, implying that the proposed framework is insensitive to the imaging errors caused by the fluoroscopic system. Instantiating partially-deployed stent segment includes mainly three steps: marker segmentation which costs 0.1s on a Nvidia Titan Xp GPU \cite{zhou2018towards_iros}, 3D marker reference prediction which costs 0.8ms, and 3D shape instantiation which costs 7ms. The total computational time is less than 0.11s, which potentially could achieve real-time running as the typical frame rate for clinical usage is around 2 $\sim$ 5 frames per second.

In the future, this paper could be combined with the 3D shape instantiation for fully-deployed \cite{zhou2018real_ral} and fully-compressed \cite{zhou2018towards_iros} stent segment to build a system of real-time 3D shape instantiation for stent grafts at any states. The Equally-weighted Focal U-Net could be retrained and integrated into the instantiation framework for improving the automation.

\section{CONCLUSIONS}
\label{sec:conclusion}
A 3D shape instantiation framework for partially-deployed stent segment was proposed in this paper, including stent segment modelling, 3D marker reference prediction, 3D marker instantiation and 3D shape instantiation. Only a single fluoroscopic image with minimal radiation is required as the intra-operative input. The Adapted GCN is introduced to explore the variation pattern of 3D markers and to provide the 3D marker references for 3D marker instantiation. Compared with the previous relevant work, the proposed framework focuses on dealing with the difficulties of predicting the stent segment shape at the partially-deployed state and achieved a comparable accuracy. 

\addtolength{\textheight}{-12cm}   




\section*{Acknowledgement}
The authors would like to thank the support of NVIDIA Corporation for the donation of the Titan Xp GPU used for this research. 


\bibliographystyle{IEEEtran}
\bibliography{reference}

\end{document}